\documentclass[10pt,twocolumn,letterpaper]{article}

\usepackage{cvpr}
\usepackage{times}
\usepackage{epsfig}
\usepackage{graphicx}
\usepackage{amsmath}
\usepackage{amssymb}

\usepackage{subcaption}

\usepackage{framed,multirow}

\usepackage{pifont}
\usepackage{graphicx}
\usepackage{array}

\usepackage{algorithm}
\usepackage[noend]{algpseudocode}

\usepackage[backend=bibtex, style=ieee]{biblatex}
\cvprfinalcopy 

\def\cvprPaperID{****} 

\begin{document}

\title{Recurrent neural networks based Indic word-wise script identification 
  using character-wise training}



\author{Rohun Tripathi\\
IIT Kanpur\\
\and
Aman Gill\\
Microsoft\\
\and
Riccha Tripati\\
IIT Delhi\\
}

\maketitle

\begin{abstract}
   This paper presents a novel methodology of Indic handwritten script recognition using Recurrent Neural Networks and addresses the problem of script recognition in poor data scenarios, such as when only character level online data is available. It is based on the hypothesis that curves of online character data comprise sufficient information for prediction at the word level. Online character data is used to train RNNs using BLSTM architecture and are used to make predictions of online word level data. These prediction results on the test set are at par with prediction results of models trained with online word data, while the training of the character level model is much less data intensive and takes much less time. Performance for binary-script models and then 5 Indic script models are reported, along with comparison with HMM models. The system is extended for offline data prediction. Raw offline data lacks the temporal information available in online data and required for prediction using models trained with online data. To overcome this, stroke recovery is implemented and the strokes are utilized for predicting using the online character level models. The performance on character and word level offline data is reported.
\end{abstract}
\section{Introduction}
\label{label:introduction}

\begin{figure}
    \begin{subfigure}[b]{0.25\textwidth}
	\includegraphics[height = 0.8in,width=1.6in]{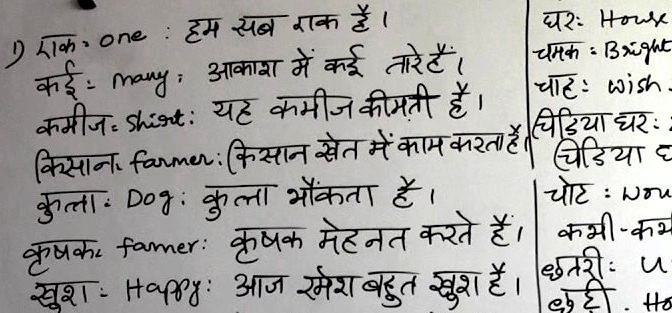}
    \end{subfigure}%
    \begin{subfigure}[b]{0.25\textwidth}
	\includegraphics[height = 0.8in,width=1.6in]{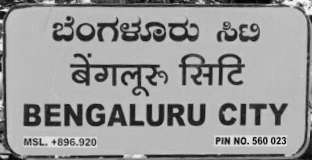}
    \end{subfigure}
  
    \caption{\textbf{Sample images with texts in multiple scripts.} The image on the left comprises handwritten Hindi and English text and the image on the right comprises printed Hindi, English and Kannada}
    \label{fig:multiScriptText}
\end{figure}

Handwritten script recognition constitutes the problem of identification of the 
script that a particular text document has been written in.
The basis for script recognition is the unique spatial relation that strokes of a particular script 
have with each other, that makes it possible to distinguish the scripts from one another. 
Most prevalent handwritten text recognition systems are language dependent, while many digital documents and images 
using multiple scripts exist, Fig.~\ref{fig:multiScriptText}, especially in geographical regions using multiple 
scripts. 
This makes handwritten script recognition an important first step towards automated 
interpretation of handwritten text documents~\cite{ghosh2010, Ubul2017ScriptIO}.

Recurrent Neural Networks (RNNs) have proven to be very effective 
~\cite{graves2, Mei2016SceneTS, Sankaran2012RecognitionOP, Ul-Hasan:2015:SLA:2880452.2880903} for handwriting 
recognition tasks. An RNN is a brand of 
Artificial Neural Networks in which connections between the hidden nodes can 
form directed loops. This architecture comprises delay functions on these 
loops enable the Neural Network to have an internal state or memory. This 
looping architecture allows RNNs to process arbitrary lengths of input and 
produce arbitrary lengths of output. For training RNNs we use rnnlib 
~\cite{rnnlib}.

To train our recognition model, we primarily use online data, which comprises the trace/strokes of a pen on a recording screen as separate characters and words are written, for character and word 
level data respectively.
Offline data comprises of images.
Online data possesses spatio-temporal information which proves to be beneficial for our script recognition task.

Existing models for script recognition focus on extracting linguistic and/or statistical models for script 
recognition at the word or the text-line level ~\cite{thaichanda, HangargeandDhandra2011, ghosh2010, Ubul2017ScriptIO}.
In contrast, our model is based on the hypothesis that the curves of online 
character level strokes of each script comprise
sufficient distinctive features for the script recognition at the word level as well. The curves are represented 
by the spatio-temporal information obtained as the text is written. Online character level training data is much 
lighter than word level data thus providing the benefit of faster training.
It further reduces the requirement of larger word level datasets. This is highly applicable in 
a country such as India, which has a diverse set of scripts, in order to bootstrap script recognition for languages where data is scarce and annotation is expensive. We illustrate by training using just the character level data and test them for word level data.

To extend our online models for prediction of offline data, we use Stroke Recovery.
A Stroke Recovery method is one which takes as input a text image and outputs 
the possible order of strokes that are required to write the character~\cite{elbaati, hamdani, kato}.
In other words, it extracts the temporal information of the handwritten data.
Training of offline data demands larger datasets and training time vis-a-vis 
that of online data.
Thus, we developed a Stroke Recovery method to obtain a possible sequence of pen strokes where each
stroke is a sequence of pixels the pen would cover.
We then classify the output into the appropriate script using recognition models 
obtained by training using the online data.

The main contributions of this paper : (i) evaluate script recognition using spatio-temporal features (ii) evaluate 
the model for 5 scripts at the character-level and apply
character-level models for word-level prediction for 4 distinct scripts, (iii) apply the same to 
offline word level recognition using stroke recovery.


\section{Recent Works}

Handwritten Script recognition~\cite{ghosh2010, Ubul2017ScriptIO} is an important component for recognition of documents written in more than a single script. Script identification in printed documents~\cite{Sankaran2012RecognitionOP} is not as complex because it does not have to learn despite the various writing styles of individuals in handwritten data. A comprehensive survey on Indic character recognition can be found in ~\cite{pal2004survey}.

A popularly explored approach to handwritten script recognition is to extract linguistic and/or statistical features 
and club it with an SVM model, ~\cite{chanda, thaichanda, ferrer}. Script identification in torn documents is addressed in ~\cite{chanda} using rotation Invariant Zernike features and 
the rotation Dependent Gradient Feature, PCA-based methods to predict orientation and then applying an SVM classifier at the character level. Following their results, recognition at the character level is more difficult when compared to script recognition at the paragraph level, line level or the word level as they use voting. However, recognition at the character level enjoys the benefit of smaller datasets and faster training times. In our approach, the system learns the curves from the character level and is applied directly on the word without a voting system. 

In ~\cite{Vinciarelli2000HMM}, a 
sliding window is applied on the image instead of segmenting the word for each character and Continuous density HMMs are trained for each letter instead of a word to 
improve generalization over unseen word. They obtain an accuracy of 83.57\% on the 
Senior and Robinson dataset. In ~\cite{Samanta2014OCR}, words  are first segmented by using a heuristic algorithm and local minimas in y-axis to generate sub-strokes. Then, they extract linear, circular features and a combination of both to train two non-homogeneous HMMs, in normal and reverse order. 

Neural networks based solutions are also popular for ­script identification, discussed in~\cite{ghosh2010}. In one of 
the earliest works, neural nets were employed for script identification in postal automation systems~\cite{Roy2005NeuralNB, Roy_2005}. 
Though MLPs are trainable, a huge amount of parameters make them harder to train. In ~\cite{Mei2016SceneTS}, a combined 
architecture using Convolutional neural networks followed by RNNs and a fully connected layer is proposed for 
script identification in images. This end-to-end architecture aims to first extract the image features, followed by an 
Bi-RNN to learn the arbitrary output for script identification and is evaluated on SIW-13 and CVSI2015~\cite{Sharma_2015_ICV}. In ~\cite{Sankaran2012RecognitionOP}, BLSTM is used for printed Devanagri Script 
recognition. For every word 5 different features are extracted (a) the lower profile, (b) the upper profile, (c) the 
ink-background transitions, (d) the number of black pixels, and (e) the span of the foreground pixels. These are then 
passed through a Bi-RNN architecture using Connectionist Temporal Classification objective function leading more the 
9\% WER improvement. In ~\cite{Ul-Hasan:2015:SLA:2880452.2880903}, a 1D-LSTM architecture, with one hidden layer is used 
for script identification at the text-line level to learn binary script models, and the prediction accuracy for 
English-Greek identification obtained is 98.19\%.
In our approach, the system is using RNNs for character level modeling and using this model for prediction at the word level. 

For offline script recognition, an approach is to extract the 
initial strokes. In~\cite{elbaati}, Elbaati et al. proposed an approach to 
stroke recovery by first segmenting the image into strokes and labeling all the edges as segments or parts of strokes. 
They run a Genetic Algorithm to optimize these strokes and an application of the above developed method is used in~\cite{hamdani}. In ~\cite{Akbari2010dictionary} authors create a dictionary of features of sub-words by using the contour information of the neighbouring sub-word in the Arabic text for the information retrieval task.
In~\cite{kato}, Kato \textit{et al.} propose a stroke recovery technique which 
works for single stroke characters. The system labels each edge in the image and bridges them in an algorithmic 
manner. Our approach for stroke recovery is similar in approach and we 
extend the system to account for multiple strokes in the input image.

\begin{figure}
    \includegraphics[width=\linewidth] {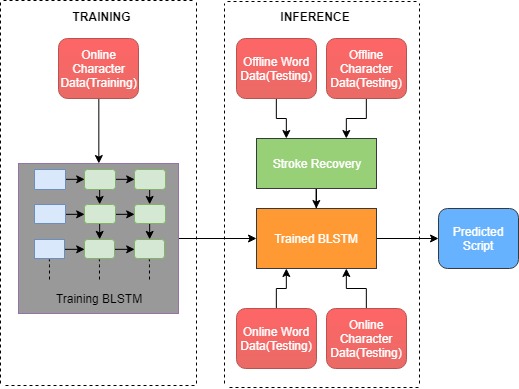}
    \caption{\textbf{Overview of the proposed method.} The RNN is trained using online character data. It can then perform at par with RNN systems trained with online word data. For offline data, stroke features are extracted and the same model is used in inference. Hence, no training data is used for offline data inference.}
    \label{fig:flow_chart}
\end{figure}

\section{Methodology}

In this section, we introduce our neural network architecture, its objective function and stroke recovery.

\subsection{RNN Model}

Our recurrent neural network uses 3-layered BLSTM architecture with Cross Entropy Error as the objective function. BLSTM comprises of Long 
short-term memory~\cite{Hochreiter:1997:LSM:1246443.1246450, graves2} layers with bidirectional connections.
The LSTM nodes have an internal architecture primarily 
comprised of three gate nodes.
At each epoch, the input gate determines the contribution of the incoming 
value to the retained value, the forget gate determines the contribution of the previous retained value 
towards the next value and the output gate controls the contribution of the retained value towards 
the output of the node. 
This allows the memory cell to preserve its state over a long range of time and 
to model the context at the feature level.

The 1D sequence recognition is improved by processing the input signal in both 
directions, i.e., one layer processes the signal in the forward direction while 
another layer processes it in the backward direction. Bi-directionality makes the network robust to erroneous strokes recovered in the backward direction by the stroke recovery methods. The output of 
both layers is combined at the next layer in a feature map.
\newline 

We used Cross Entropy Error as our objective function. In a binary network, the error $E$ is:

\begin{equation}
  E = - \sum_{z \epsilon S} z\ln y + (1-z)\ln(1-y)
  \label{eq:binary}
\end{equation}

Where $z$ is the target class (0 or 1), and $y$ can be understood as the 
probability that the input belongs to a class. Details can be found in  ~\cite{graves2}.

When extended to multi-class networks with k classes, the error function $E$ becomes:

\begin{equation}
  E = - \sum_{z \epsilon S} \sum_{k=1}^{K} z_{k}\ln y_{k}
\end{equation}

Similarly, Extending (2) for multiple classes.

\begin{equation}
  p(z|x) = \prod_{k=1}^{K} y_{k}^{z_{k}}
\end{equation}

\subsection{Offline to online conversion - Stroke Recovery}

\begin{figure}
    \begin{subfigure}[b]{0.25\textwidth}
      \includegraphics[height = 1.0in,width=\linewidth]{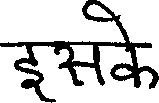}
      \caption{Initial Image}
      \label{fig:stroke_recov_init}
    \end{subfigure}%
    \begin{subfigure}[b]{0.25\textwidth}
      \includegraphics[height = 1.0in,width=\linewidth]{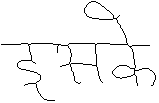}
      \caption{Skeleton}
      \label{fig:stroke_recov_skeleton}
    \end{subfigure}\\
    \begin{subfigure}[b]{0.25\textwidth}
      \includegraphics[height = 1.0in, width=\linewidth]{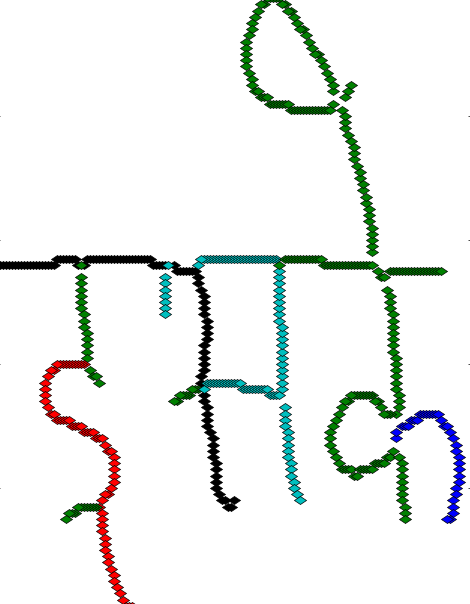}
      \caption{Recovered Strokes}
      \label{fig:stroke_recov_recov}
    \end{subfigure}%
    \begin{subfigure}[b]{0.25\textwidth}
      \includegraphics[height = 1.0in, width=\linewidth]{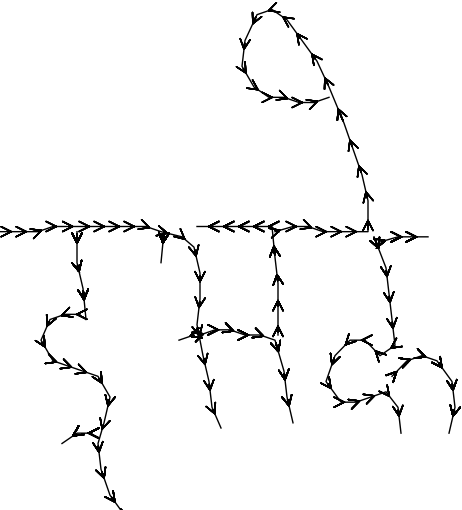}
      \caption{Stroke directions}
      \label{fig:stroke_recov_direction}
    \end{subfigure}%
    \caption{\textbf{Key steps in Stroke recovery process} (a) Sample of initial raw image, (b) Skeleton extracted from the raw image using Fiji, (c) Independent strokes extracted by the system depicted by individual colors, (d) Directions corresponding to each of the strokes retrieved.}
    
	\label{fig:stroke_recov}
\end{figure}

\begin{figure}
    \begin{subfigure}[b]{0.20\textwidth}
      \centering
      \includegraphics[height = 1in,width=0.9in]{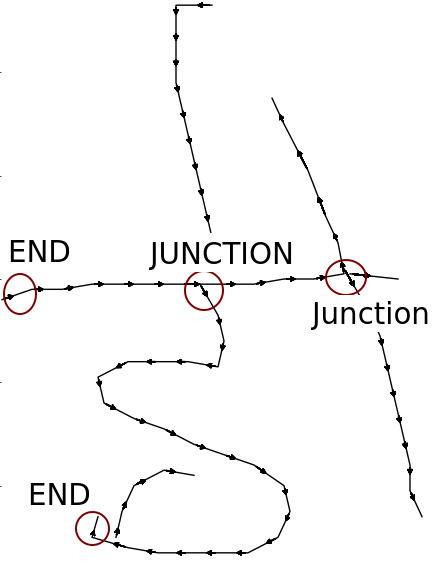}
      \caption{Critical Points}
      \label{fig:critical_points_D}
    \end{subfigure}\hspace{.5cm}
    \begin{subfigure}[b]{0.20\textwidth}
      \centering
      \includegraphics[height = 1in,width=0.9in]{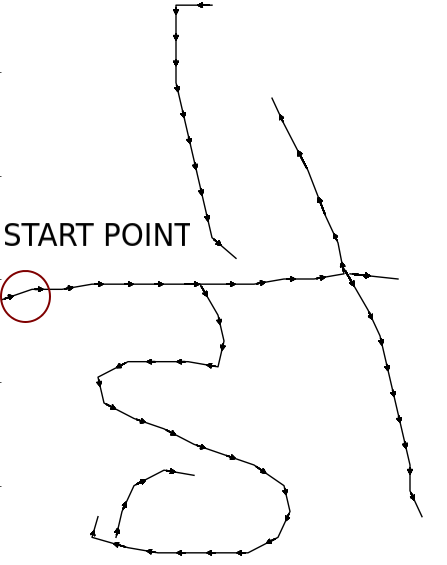}
      \caption{Start Point Selection}
      \label{fig:start_point_selection}
    \end{subfigure} \hspace{.5cm}\\
    \begin{subfigure}[b]{0.20\textwidth}
      \centering
      \includegraphics[height = 1in,width=0.9in]{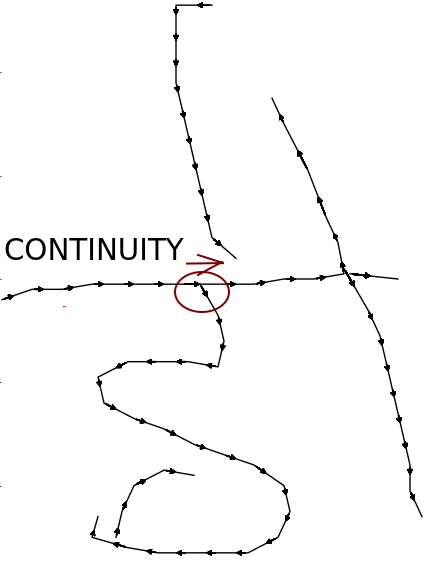}
      \caption{Outlet Selection}
      \label{fig:method_continuity}
	\end{subfigure}\hspace{.5cm}
    \begin{subfigure}[b]{0.20\textwidth}
      \centering
      \includegraphics[height = 1in,width=0.9in]{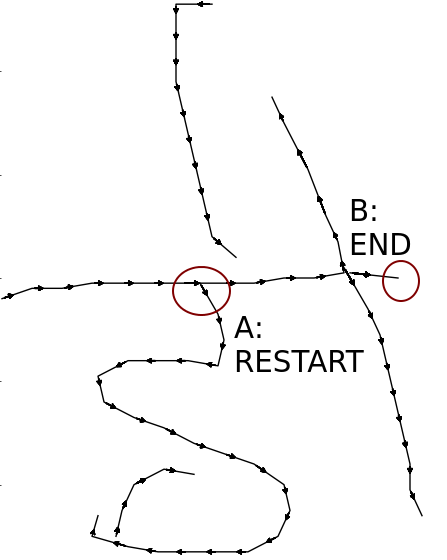}
      \caption{Restart Selection}
      \label{fig:selectingStartPointAgain}
    \end{subfigure}
    \caption{\textbf{Each of the Stroke recovery strategies.} (a) Critical 
	points (junction and end points) are marked, (b) Initial Start point is the point nearest to top-left corner, (c) outlet stroke from a junction selected based on preserving continuity, (d) Re-selection of start point in case of multiple strokes}  
\end{figure}

To utilize neural networks trained using online character level data for offline 
data prediction, we extract the the temporal data using stroke recovery, Fig.~\ref{fig:stroke_recov_direction}.
The first step of stroke recovery is to obtain the skeleton of the provided binary image dataset 
for which we used Fiji's ~\cite{fiji} skeletonize functionality, Fig.~\ref{fig:stroke_recov_skeleton}.
The system next calculates the critical points, Fig.~\ref{fig:critical_points_D}, which 
comprise the endpoints, points connected to only one other point, and the 
junction points, points connected to more than two points. Junction points lie on at least two strokes in a given image.

\begin{figure}[h]
  \centering
  \includegraphics[width=1.5in]{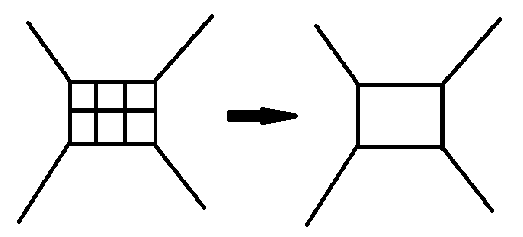}
  \caption{\textbf{Pictorial Depiction of conjunction of junction points} Neighboring cluster of junction points are merged into a joint junction area.}            
  \label{fig:convertJPs}
\end{figure}

The method then reduces each junction point and its neighboring junction points 
into one joint junction area, by connecting each junction point in a 
neighborhood with each of the outlets of each of the neighboring Junction points Fig.~\ref{fig:convertJPs}.

\begin{algorithm}
\caption{Merging of Junction points (Conjunction)}\label{alg:euclid}
\begin{algorithmic}[1]
\Procedure{Conjugate}{$junctionPts$}
\ForAll{junction pixel i in junctionPts}
\State $neighbours\gets getNeighbouringPixels(i)$
\ForAll {junction pixel j in neighbours}
\ForAll {pixel k in j's neighbours}
\If {k is not a junction pixel}
\State AddPixelToJunctionBranch(k, i)
\EndIf
\EndFor
\EndFor
\EndFor
\EndProcedure
\end{algorithmic}
\end{algorithm}

To begin stroke recovery, our method needs to select a suitable start point and 
it utilizes a straight-forward strategy to select the endpoint closest to the 
top most corner of the image as the start point. This is based on the insight 
that the scripts we are working on are written in the left to right order and  
the strokes begin from the top most corner, Fig.~\ref{fig:start_point_selection}.
To justify this selection objectively, we took 1000 ground truth online character level data points for English and Hindi. For each each image, we calculated the start and end points of each stroke and the stroke that will be selected using the method we propose. if our method indeed selects a start point and not an end point, it would count as a correct selection. We had 957 and 903 correct selections for English and Hindi, respectively. Demerits of this strategy are discussed in the results under error analysis.

On reaching a junction area during the stroke recovery process, the system 
calculates the slope of the incoming curve and the slope of each of the exiting 
curves from the junction area. The slope 
based selection method selects the outgoing curve that has the slope closest to 
that of the incoming curve Fig.~\ref{fig:method_continuity}. This strategy is based on maintaining continuity 
and avoiding jerks to recover the most probable path of the stroke.

In case the image contains multiple and disjoint strokes, we might reach an 
endpoint in the stroke recovery method before all the points have been covered.
This implies that all reachable points have been covered and prompts a need to 
select the next start point to continue stroke recovery and cover the rest of 
the strokes.

The selection of the next start point needs to be made from the set of all 
calculated endpoints and the neighbors of the junction points encountered thus 
far.
We experimented with two strategies here, the first is to prioritize selection 
from the end points over the neighbors of the junction points. Further, among 
the endpoints, the ones closest to the top left corner are given higher 
priority.
The second approach is to prioritize the neighbors of the junction points over 
the end points the priority, where among the neighbors, the priority is given in 
a first come first serve basis, Fig.~\ref{fig:selectingStartPointAgain}. The latter approach outperforms the former in 
terms of accuracy of predictions made on the strokes retrieved.
\section{Datasets and Experiments}
This section covers the datasets, training details and approach for online and offline data.
\subsection{Datasets}
The sources for online data are as follows. We used IAM Online ~\cite{iamol} for the English language word level data and Chars74K dataset~\cite{chars74k} for English character level data.
For Hindi, Tamil and Telugu languages, the online character and word data is obtained from the Lipi Toolkit, HP 
Labs~\cite{Madhvanath:2007:LipiTk}. For the Bangla script, the online character and word data is obtained from 
CMATERdb~\cite{cmaterdb} and the Assamese online character data is obtained from the UCI Repository~\cite{uciuji}.

The sources for offline data are as follows. We used IAM Dataset ~\cite{iamword} for the English language word (line) level data and Chars74K dataset~\cite{chars74k} for the character level data.
For Hindi and Bangla, the offline character and word data is obtained from ~\cite{cmaterdb}.

\subsection{Training Details}
For training our RNN with Cross-Entropy loss, we used gradient descent optimizer with momentum set at 0.9 and a learning rate of $1e^{-4}$. The RNN model architectures used 3 BLSTM layers. The training was stopped when the result on the validation data did not improve for  20 epochs.

Two RNN instances trained using the same data might converge to different local minimum. A combination model approach attempted to club the results of different models to lead to an improvement in overall prediction.
For a test word, the model takes the predictions by different networks and the cross entropy error for each prediction. If the predictions mismatch, it computes the error which indicates the confidence of the prediction. For a binary model, the error for prediction Class $C_{0}$ when error for $C_{1}$ is present can be calculated using  \ref{eq:binary}.

\begin{equation}
  E_{0} = - \ln(1-e^{-1*E_{1}})
\end{equation}

The model chooses the prediction with the lower error value as the final output. This method was used to qualitatively improve predictions made by the networks on word level data recovered by stroke recovery.
For encoding a data point comprising a sequence of strokes, we appended the data together, as a series of vectors $v$, each vector with three values.
The first and the second values are the x and y coordinate of each pixel 
in the stroke and the third value is 1 or 0, depending on whether the pixel is 
the first pixel of a stroke, or not. The datasets were obtained from various 
different sources and recording devices.
This prompted a need for a universal Normalization Technique for the vectors in 
the training data, to eliminate features specific to the dataset and retain only 
features of the script. The following standardization is used:

\begin{equation}
    x^{(i)} = \frac{ v^{(i)}-\mu}{\sigma}
\end{equation}

Where vector, $v$ is the array of three values and $\mu$ and $\sigma$ are the 
mean and the standard deviation of all the vectors in the particular dataset.

\subsection{Online data approach}
\label{label:online_data_approach}
Our proposed model, Fig.~\ref{fig:flow_chart} is based on online character level data training and results are reported in \ref{table:online}. We first report results of the model trained for two languages, Hindi and English. The input provided had 12500 instances of online character level data from each of the languages and the model  converged in 116 minutes. The test set comprised of 2000 data points from each class.  The script recognition of the word level handwritten online data was then tested  on the above binary character level model. The test set selected comprised of 2250 test cases of English and Hindi. Subsequently, the system was extended to 5 languages by adding the datasets for Bengali, Tamil and Telugu languages to the training.

\subsubsection{Baselines for online data models}
The first baseline for this setting is models trained with online character level data using other sequence learning models HMMs and Structured Perceptrons. 
HMM and Structured Perceptron models are trained with the same data as the RNN models. In the HMM model, an individual HMM was trained for each label class at training time using hmmlearn Toolkit. During inference, the probability of the sequence belonging each of the trained classes is calculated and the label with the highest probability is assigned to the test sequence. HMMs converge to a local solution depending on a random initial state and we found the optimal combinations of HMM models of each language by calculating the result of different combinations of models on a validation set. The Structured 
Perceptron is discussed in~\cite{Collins:2002:seqlearn} and we use its implementation from the seqlearn Toolkit. The performance comparison on character and word level test data is presented in Table \ref{table:online}. 

Our overall motivation if to prove that models learned using the character level data perform at par with models trained with word level data, based on the hypothesis stated in \ref{label:introduction}. Hence, the second baseline to verify against is sequence models trained using word level data. 
To compare models trained using character level data, we trained models using online word English and Hindi data, which required a considerably longer training time, about 8.3 times longer than models trained with character level data. Thereafter, we trained word level models using four Indic scripts, Hindi, Bengali, Telugu and Tamil and one Latin script, English, using 2000 samples from each of the languages. We tested the prediction for word level data with 2000 instances from four languages as the Telugu on-line word dataset was not available. Results with these baseline models are detailed in Table \ref{table:online}.

As a third baseline for our online data based models, we use the results from \cite{Ul-Hasan:2015:SLA:2880452.2880903}. It must be noted that they report binary class script recognition results for English and Greek scipts, while we report results for English and Indic scripts. All results are reported in Table \ref{table:online}. 

\begin{table}[t]
  \caption{Results on Test sets of online data}
  \small
  \begin{tabular}{|c|c|c|}
    \hline	
    Eng-Hindi Model	& Test-Char level &Test-Word Level \\ \hline		
    \textbf{RNN Char Model }            & \textbf{99.75\% }                             			
    &\textbf{ 100\%}\\
    RNN Word Model             & -                                  			
    & 100\% \\ 
    HMM Char Model             & 95.70\%                                			
    & 77.66\%\\
    HMM Word Model             & -                                  			
    & 98.60\% \\ 	 
    Perceptron Char Model             & 77.38\%                                  			
    & 56.6\% \\ 	 
    Perceptron Word Model             & -                                  			
    & 94.10\% \\ 	 
    Baseline Eng-Grk (\cite{Ul-Hasan:2015:SLA:2880452.2880903})             & -                         			
    & 98.19\% \\ 	 
    \hline
    Many-language Model                   & 5-Lang Char 				
    &4-Lang Word \\ \hline
    \textbf{5-Lang RNN Char Model} 		& \textbf{99.74\% }                             
    &\textbf{ 99.82\% }           \\
    4-Lang RNN Char Model     	& -                                      
    & 99.7\% \\  \hline
  \end{tabular}
  \label{table:online}
\end{table}

\subsubsection{Comparison of training of character and word level models}
We analyze the training time and data consumption of character data versus word level data. We find that the average number of strokes in the English and the Hindi character dataset is 1.9 and 3.7, whereas the average number of strokes in the English and the Hindi word dataset is 11.2 and 13.2. The average time taken to train the model using a single stroke varies based on batching, but when using the exact same batching technique, time taken by word level models to converge is 8.3 times greater than time taken by character level models to converge. This gain linearly translates to gain in data storage and annotation time needed when using character level over word level data.


\begin{table}[]
  \centering
  \caption{Results on offline character and word test sets.}
  \begin{tabular}{|c|c|}
    \hline
    Recovered Strokes - Char Data & Prediction - online Model\\\hline
    \textbf{Eng-Hin RNN}                                                    & \textbf{89.2\%}  \\
    Eng-Ban RNN                                                    & 86.2\%   
    \\
    Eng-Tam RNN                                                    & 87.6\%  \\
    Eng-Hin HMM                                                    & 68.3\%  \\
    Eng-Hin-Tam-Ban raw pixel                                      & 80.8\%  \\
    \hline
    
    Recovered Strokes - Word Data & Prediction - online Model\\\hline
    \textbf{Eng-Hin RNN}                                                     & \textbf{75.6\%}   \\
    Eng-Ban RNN                                                    & 72.9\%  \\
    Eng-Hin RNN Clubbed                                               & 75.2\% \\ 
    Eng-Hin HMM                                                     & 52.3\%  \\
\hline
  \end{tabular}
  \label{table:offline}
\end{table}

\subsection{Offline data approach}
We retrieved strokes from the character and word images using our stroke 
recovery method. All results reported in this section are prediction results on networks trained with online character level data as described in \ref{label:online_data_approach}. Our experiments were restricted to binary classification models for English and one Indic script, one of Hindi, Bangla or Tamil for character. For word level classification, we did not have Tamil data. For character and word level data, we extract strokes for 250 character images from each of the languages. The results for are reported in \ref{table:offline}.

\subsubsection{Baselines for offline data models}
The first baseline for our offline approach is using HMM models trained with online character level data as described in \ref{label:online_data_approach}.

The second baseline is using the raw pixel values of the offline images as sequences. We trained for Hindi, Bengali, Tamil and English using 550 instances of each language. Different datasets present images with varying size. We normalized each image by resizing the height of the character image to a constant value, 400 pixels, while retaining the aspect ratio of the original image. We chose 400 pixels as it is smaller than the average height from each language dataset. We calculate the new width as follows:

\begin{equation}
width = originalWidth*\frac{400}{originalHeight}
\end{equation}

Images from different datasets have different stroke widths. This variation was removed by thinning the image to a unit width image followed by thickening the stokes to a uniform width, using Fiji ~\cite{fiji}. Using raw pixel values is a data intensive method and hence this baseline is limited to character level. Results of all baselines are reported in \ref{table:offline}.

\begin{table}[h!]
  \centering
  \caption{Qualitative Results, Character and Word Level Offline data.}
  \begin{tabular}{ | c | m{1.5cm} | m{1.5cm} | m{1.8cm} | }
    \hline
    \centering
    
    Images & Offline Model & Online Model & Combination of Models\\ \hline
    
    \begin{minipage}{.04\textwidth}
      \includegraphics[width=\linewidth, height=6mm]{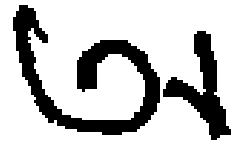}
    \end{minipage} & \ding{51} & \ding{51}& -\\ 
    
    \begin{minipage}{.04\textwidth}
      \includegraphics[width=\linewidth, height=6mm]{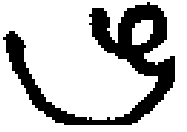}
    \end{minipage} & \ding{55} & \ding{51}& -\\ 
    
    \begin{minipage}{.08\textwidth}
      \includegraphics[width=\linewidth, height=6mm]{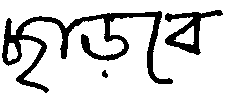}
    \end{minipage} & - & \ding{55}& -\\ 
    
    \begin{minipage}{.08\textwidth}
      \includegraphics[width=\linewidth, height=6mm]{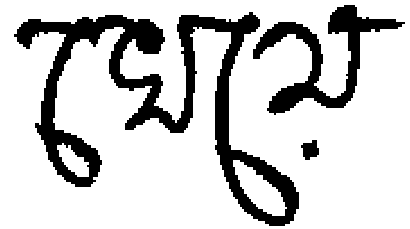}
    \end{minipage} & - & \ding{51}& -\\ \hline
    
    \begin{minipage}{.04\textwidth}
      \includegraphics[width=\linewidth, height=6mm]{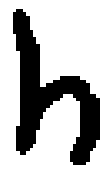}
    \end{minipage} & \ding{51} & \ding{51} & \ding{51}\\ 
    
    \begin{minipage}{.04\textwidth}
      \includegraphics[width=\linewidth, height=6mm]{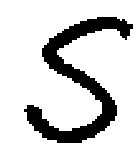}
    \end{minipage} & \ding{51} & \ding{51} & \ding{51}\\ 
    
    \begin{minipage}{.06\textwidth}
      \includegraphics[width=\linewidth, height=6mm]{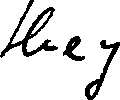}
    \end{minipage} & - & \ding{51} & \ding{51}\\
    
    \begin{minipage}{.06\textwidth}
      \includegraphics[width=\linewidth, height=6mm]{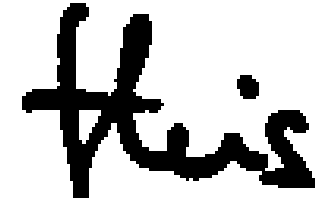}
    \end{minipage} & - & \ding{55} & \ding{51}\\ \hline
    
    \begin{minipage}{.04\textwidth}
      \includegraphics[width=\linewidth, height=6mm]{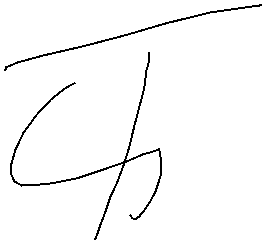}
    \end{minipage} & \ding{51} & \ding{51} & \ding{51}\\ 
    
    \begin{minipage}{.04\textwidth}
      \includegraphics[width=\linewidth, height=6mm]{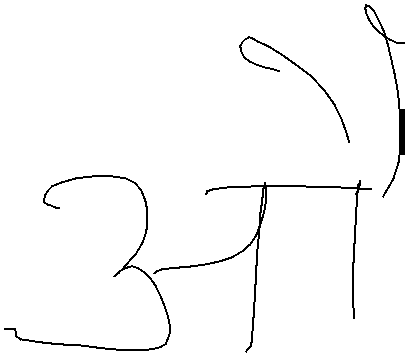}
    \end{minipage} & \ding{51} & \ding{51}  & \ding{51} \vspace{0.3cm}\\ 
    
    \begin{minipage}{.09\textwidth}
      \includegraphics[width=\linewidth, height=6mm]{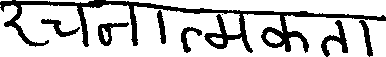}
    \end{minipage} & - & \ding{55} & \ding{51}\\ 
    
    \begin{minipage}{.08\textwidth}
      \includegraphics[width=\linewidth, height=6mm]{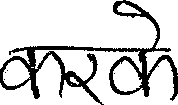}
    \end{minipage} & - & \ding{51} & \ding{51}\\
    
    \hline
  \end{tabular}
  \label{table:qualitative}
\end{table}

\begin{figure}
  \begin{subfigure}[b]{0.25\textwidth}
    \centering
    \includegraphics[width=1.2in, height = 1.3in]{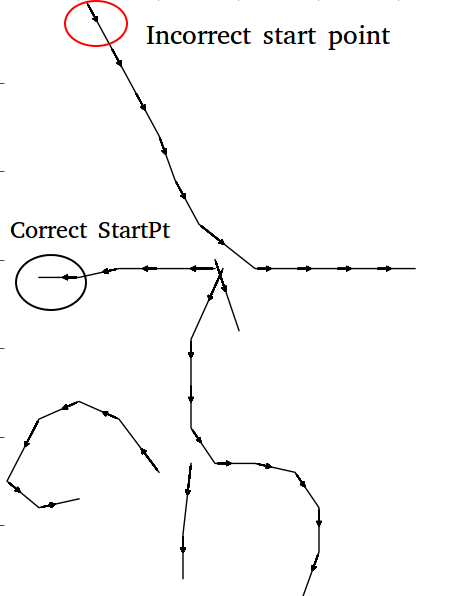}
    \caption{}
  \end{subfigure}%
  \begin{subfigure}[b]{0.25\textwidth}
    \centering
    \includegraphics[width=1.2in, height = 1.3in]{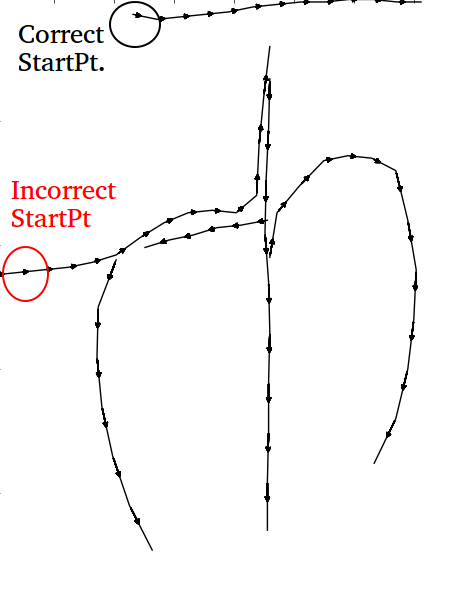}
  \end{subfigure}
  \caption{\textbf{Errors in start point selection.} The start point selection might select an incorrect point at times as indication with the red outline. The black outline is the ground truth value.}
    \label{fig:starterror}
\end{figure}
  
\begin{figure}
    \begin{subfigure}[b]{0.15\textwidth}
      \includegraphics[width=0.9in]{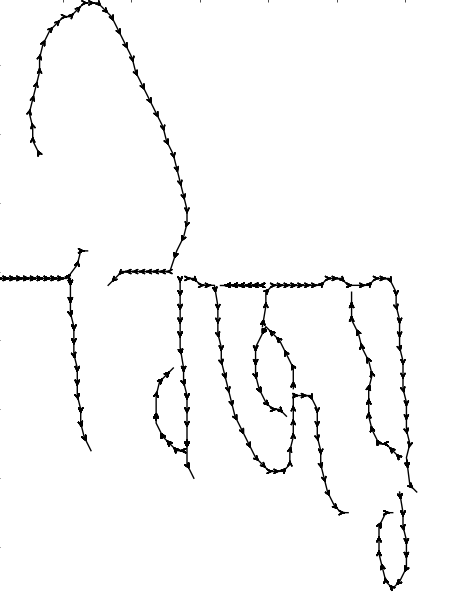}
      \caption{Bangla}
    \end{subfigure}%
    \begin{subfigure}[b]{0.15\textwidth}
      \includegraphics[width=0.9in]{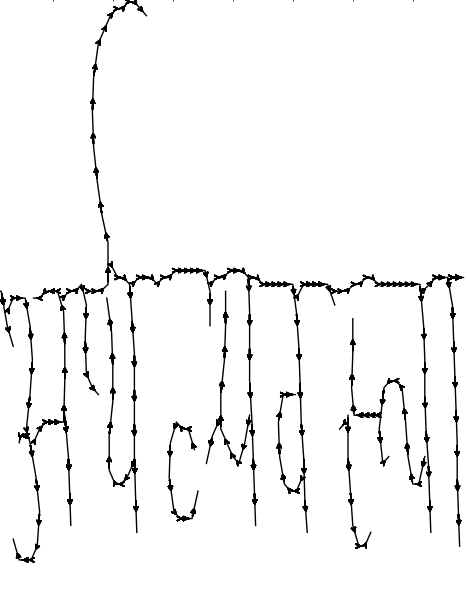}
      \caption{Hindi}
    \end{subfigure}%
    \begin{subfigure}[b]{0.15\textwidth}
      \includegraphics[width=.9in]{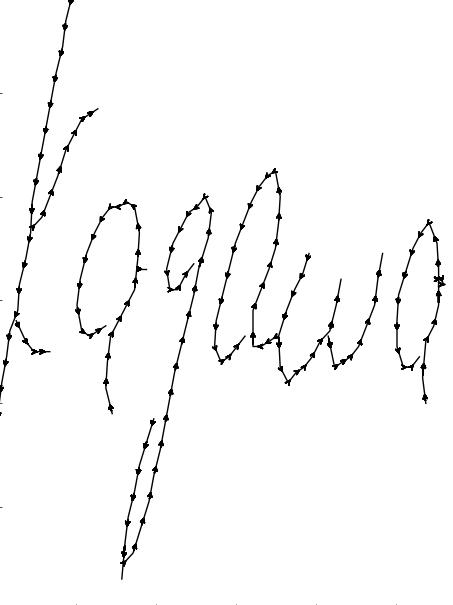}
      \caption{English}
    \end{subfigure}
    \caption{\small{\textbf{Qualitative Results} Correctly classified complex samples} }
    \label{figure:qualitative_long}
\end{figure}

\section{Results}

Our RNN model, trained on character-level data for two classes, English and Hindi, achieved a 99.75\% accuracy on the test character data and a 100\% accuracy on the test word data, Table \ref{table:online}.
Hence, our approach shows that one can train script classifiers using character level online data that for script recognition of online word level data. This demonstrates our basic hypothesis that the curves, or the temporal data, of online character level data comprise sufficient features for script recognition task of word level data and script recognition is not a word level task. Our RNN model also considerably outperforms the baseline character level HMM and Structured Perceptron models.

These results hold when multi-class character level models are used for multi-class word level classification. The character level model achieved an accuracy of 99.82\% on the test word data of four classes, while the word level model achieved a marginally lower accuracy of 99.7\%. Hence, one can train script classifiers on character level data in scarce data settings, in settings in which annotation of data is expensive, unavailable and in settings where training data and time has a significant cost.

We demonstrate the applicability of the our online character level models for recognition of offline character and word level data, which lack temporal data, by recovering the original strokes. The recovered strokes are classified using models trained on online data, Table \ref{table:offline}. The best result is achieved for binary script classification for English and Hindi offline character data, with an accuracy of 89.2\%. For word level data, English and Hindi offline script recognition achieves the best accuracy at 75.6\%. Hence, we demonstrate how online data can be used for classification.

\subsection{Model Combination}

We experimented with a combination model approach which clubs the results of different 
RNN models to provide a final output. The system was provided with the model predictions of two separate networks for English-Hindi offline word data with initial prediction.
The clubbed model got an accuracy of 75.2\%, with improved recognition of instances that are visually easier to disambiguate, Table \ref{table:qualitative}.

\subsection{Error Analysis}
A major discrepancy is seen in the results between the predictions for online and offline data, both at the character and at the word level. A possible is in the stroke recovery method.
The method uses heuristics for tasks such as selection of a start point for the stroke recovery which might 
be erroneous at times, Fig.~\ref{fig:starterror}.
Also, the continuity model for outlet selection can be further improved by using contour information of the incoming curve and outgoing paths at a junction. The second major discrepancy is seen in the predictions for the offline character level data and the offline word level data. Any error in the stroke recovery of an image propagates and leads to a cascading effect in the subsequent strokes. Since the average number of strokes in a word is greater than in a character, about 6 times in English and about 4.5 times in Hindi, this error builds up. A possible solution to this could be to divide the word 
into histograms and apply stroke recovery individually on each of the regions, followed by prediction of the individual entities and selection via a voting system. A future direction of this work to reduce the discrepancy 
between the online and offline results is to combine online and offline 
features for better character wise training.

\section{Conclusion}

This paper proposes a unique approach for word-wise script 
recognition using character-wise training of Recurrent Neural Networks.
This is demonstrated for both online and offline datasets.
It demonstrates the basic hypothesis that the curves, or the temporal data, 
of online character level data comprise sufficient features 
for script recognition task of word level data. Training 
networks at the character level has major benefits such as faster training and requires marginal
data, vis-a-vis word-level training. Thus, it is highly applicable to bootstrap 
script recognition using limited data for a diverse set of scripts and stands out from previous models that focus on 
extracting linguistic and/or statistical models and require more data for script recognition.

Offline data lacks the temporal data, which forms a crucial part for script 
prediction using our model. To over come this, we have developed a 
stroke recovery system for retrieving the strokes for offline character level 
and word level data and then using networks trained with online character level data for script prediction.

\newpage


\printbibliography

\end{document}